\documentclass[11pt]{article}
\usepackage{amsmath}
\usepackage[final]{acl}

\usepackage{times}
\usepackage{latexsym}
\usepackage{subcaption}
\usepackage[normalem]{ulem} % provides \sout
\usepackage{xcolor}
\usepackage{booktabs}

\usepackage[T1]{fontenc}
\usepackage[utf8]{inputenc}

\usepackage{microtype}

\usepackage{inconsolata}

\usepackage{graphicx}

\usepackage{tcolorbox}

\usepackage{multirow}

\title{A Shared Geometry of Difficulty in Multilingual Language Models}
\author{
Stefano Civelli\textsuperscript{1}, 
Pietro Bernardelle\textsuperscript{1}, 
Nicol\`o Brunello\textsuperscript{2}, 
Gianluca Demartini\textsuperscript{1}
\vspace{0.2em}
\\
\textsuperscript{1}The University of Queensland \quad
\textsuperscript{2}Polytechnic University of Milan
\\
\small\texttt{\{s.civelli,p.bernardelle,g.demartini\}@uq.edu.au, nicolo.brunello@polimi.it}
}

\begin{document}
\maketitle
\begin{abstract}
Predicting problem-difficulty in large language models (LLMs) refers to estimating how difficult a task is according to the model itself, typically by training linear probes on its internal representations.
In this work, we study the multilingual geometry of problem-difficulty in LLMs by training linear probes using the AMC subset of the Easy2Hard benchmark, translated into 21 languages.
We found that difficulty-related signals emerge at two distinct stages of the model internals, corresponding to \textbf{shallow} (early-layers) and \textbf{deep} (later-layers) internal representations, that exhibit functionally different behaviors. Probes trained on deep representations achieve high accuracy when evaluated on the same language but exhibit poor cross-lingual generalization. In contrast, probes trained on shallow representations generalize substantially better across languages, despite achieving lower within-language performance.
Together, these results suggest that LLMs first form a language-agnostic representation of problem difficulty, which subsequently becomes language-specific. This closely aligns with existing findings in LLM interpretability showing that models tend to operate in an abstract conceptual space before producing language-specific outputs. We demonstrate that this two-stage representational process extends beyond semantic content to high-level meta-cognitive properties such as problem-difficulty estimation.
\end{abstract}

\section{Introduction}
Large language models (LLMs) are increasingly deployed in multilingual settings, yet our understanding of their internal reasoning remains heavily skewed toward English \cite{resck-etal-2025-explainability}. While recent work suggests that LLMs may internally ``think'' in English or exhibit an English-centric representational topology \cite{chang2022geometry,kim2025language,li2025exploring,schut2025multilingual,wendler2024llamas}, less is known about whether higher-level meta-cognitive attributes (e.g., the model’s internal estimate of how difficult a problem is) generalize across languages.

\citet{lugoloobi2025llms} showed that problem difficulty can be decoded linearly from model residual activations. Focusing on English inputs, they demonstrated that these difficulty signals enable effective interventions, reducing hallucinations via “difficulty vectors” steering and serving as a robust proxy for performance generalization during reinforcement learning.

\begin{figure}
    \centering
    \includegraphics[width=1\linewidth]{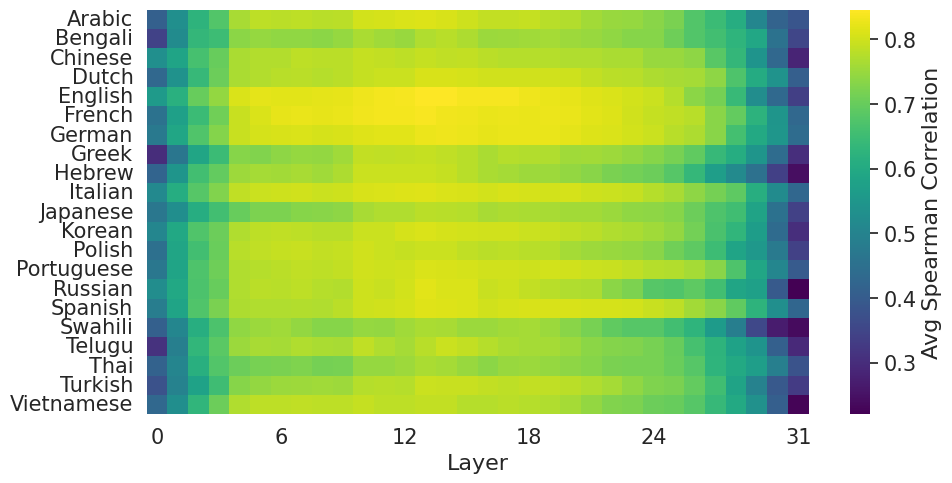}
    \caption{\textbf{Layer-wise performance of difficulty probes across languages for \texttt{LLaMA-3.1-8B}.} 
    Heatmap shows, for each test language (rows) and transformer layer (columns), the average Spearman correlation between probe predictions and ground-truth difficulty, where each value is averaged over probes trained on all other languages. Performance peaks in the middle layers, indicating that difficulty representations are most consistently aligned across languages at intermediate depths of the network.\vspace{-0.5em}}
    \label{fig:layerwise-transfer}
\end{figure}

Whether such internally encoded notions of difficulty persist beyond English, however, remains unexplored.  If a model encounters the same mathematical problem expressed in different languages, does it construct distinct difficulty representations, or does it project all inputs onto a shared notion of “difficulty” in activation space?
In this work, we address this gap by bridging difficulty probing with multilingual representation learning. We construct a multilingual version of the American Mathematics Competitions (AMC) dataset spanning 21 languages and train linear probes to predict problem difficulty across languages and model layers. We found that problem difficulty is encoded in a depth-dependent manner that trades off cross-lingual generalization and language-specific accuracy.
Linear probes show that difficulty is decodable both from shallow and deep layers, but with distinct behavior: shallow layers support strong cross-lingual transfer (see Figure~\ref{fig:layerwise-transfer}), while deeper layers yield higher accuracy within the training language but generalize poorly across languages, indicating an early language-agnostic representation that is later refined into language-specific form.

\section{Related Work}

\paragraph{Difficulty Encoding in LLMs.}
Recent work shows that LLMs encode high-level meta-properties of tasks in their internal representations. Most notably, \citet{lugoloobi2025llms} demonstrate that \emph{human-perceived problem difficulty} is strongly linearly decodable from residual activations across models and domains using Easy2Hard-Bench \cite{ding2024easy2hard}. They further validate the utility of this feature, showing that manipulating the difficulty direction (steering) can suppress hallucination and improve responses. However, their analysis is restricted to English inputs.

\paragraph{Multilingual Internal Representations.}
Several studies suggest that multilingual LLMs do not maintain fully language-agnostic internal spaces. \citet{schut2025multilingual} show that multilingual models perform key reasoning steps in representations closest to English, even when operating in other languages. Similarly, \citet{li2025exploring} find that probing performance degrades substantially for low-resource languages and that deeper layers become increasingly language-specific, with reduced cross-lingual representational similarity. These results connect to broader analyses of cross-lingual alignment and multilingual geometry, which find persistent language-sensitive directions alongside language-neutral structure in representation space \citep{chang2022geometry,hammerl-etal-2024-understanding,kim2025language}.

\vspace{0.5em}
We bridge these lines of work by studying difficulty as a multilingual internal signal. While prior work has examined difficulty encoding in monolingual settings and language dependence in multilingual models largely through linguistic or reasoning probes, we connect these strands by framing difficulty as a cross-lingual internal property.

\section{Methodology}

\subsection{Data}
\label{sec:data}

We use the AMC subset of the Easy2Hard benchmark \cite{ding2024easy2hard}, which contains approximately 4,000 math problems annotated with continuous difficulty scores.
Difficulty is estimated via Item Response Theory (IRT) from human success rates, yielding values in $[0,1]$.

We translated the original English problems into 20 additional languages using \texttt{gpt-5.1}~\cite{OpenAI_GPT5-1_2026} resulting in a 21-language benchmark (costs in Appendix \ref{a:costs}). For each language, we use approximately 3{,}000 problems for training and 1{,}000 for testing.
The train–test split is defined once at the level of problem, ensuring that cross-lingual evaluation is performed on identical unseen problems (discussion on translation quality in Appendix \ref{a:t_quality}).

\begin{figure*}[t!]
    \centering
    \includegraphics[width=0.5\linewidth]{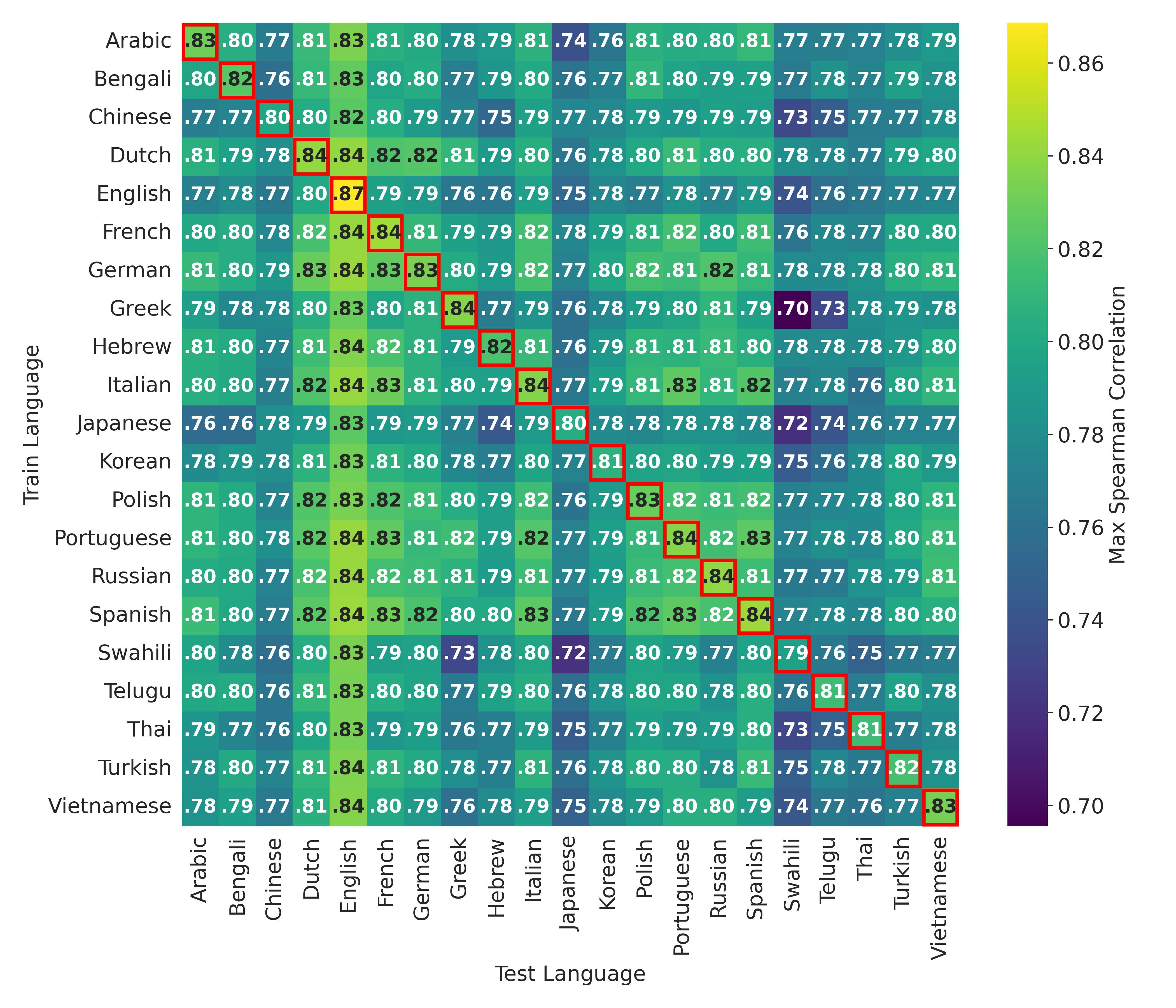}\hfill
    \includegraphics[width=0.5\linewidth]{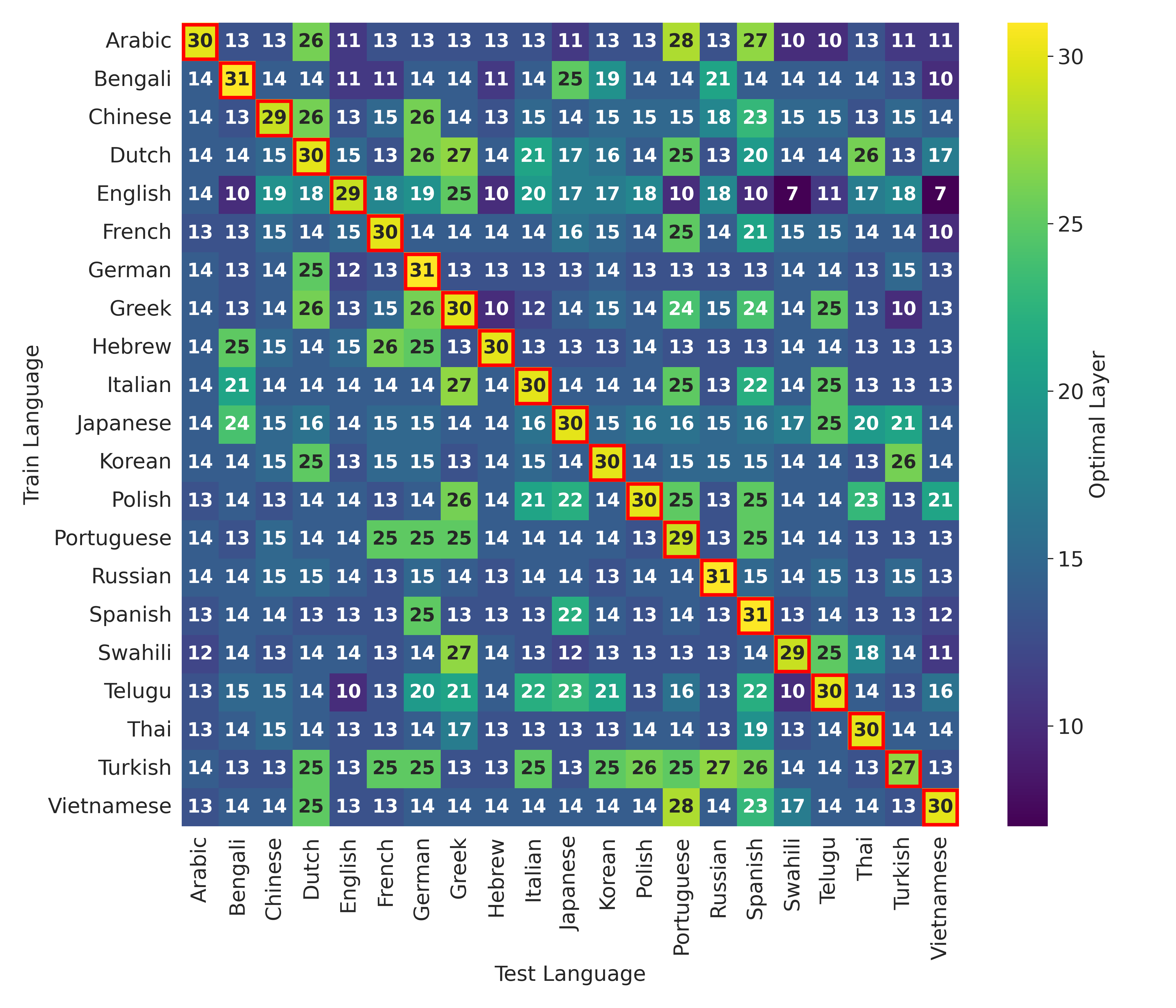}
    \caption{\textbf{Cross-lingual structure of difficulty representations in \texttt{LLaMA-3.1-8B}.}
    \textbf{Left:} Maximum Spearman $\rho$ achieved by linear difficulty probes for each training--testing language pair, evaluated at the layer that maximizes performance for that pair. Diagonal entries correspond to same-language probing, while off-diagonal entries reflect cross-lingual transfer.
    \textbf{Right:} Transformer layer indices at which peak performance is attained for each language pair. }
    \label{fig:spearman_matrices_Llama3_1}
\end{figure*}

\subsection{Models}
\label{sec:models}

We evaluate four instruction-tuned LLMs spanning different architectures and scales: \texttt{LLaMA-3.1-8B} \cite{llama3_1}, \texttt{LLaMA-3.2-3B} \cite{llama3_2_1b_3b}, \texttt{LLaMA-3.2-1B} \cite{llama3_2_1b_3b} and \texttt{Qwen3-8B} \cite{qwen3}.
All models are prompted using their standard chat templates to reflect realistic deployment conditions.
To ensure basic linguistic adequacy, we qualitatively verify that each model preserves meaning when translating a small subset of prompts from each language back into English.

\subsection{Experimental Setup}
\label{sec:exp}

\paragraph{Feature Extraction.}
Using TransformerLens \cite{nanda2022transformerlens}, we extract residual stream activations from every transformer layer.
For each input, we record the residual vector at the final prompt token, as this token has been shown to provide the most informative representation for linear probing of problem difficulty \citep{lugoloobi2025llms}.

\paragraph{Probing and Evaluation.}
We train linear Ridge regression probes to predict continuous difficulty scores from residual activations.
Probes are trained independently at each layer to analyze the depth-wise emergence of difficulty representations.
Each probe is trained on a single language and evaluated either monolingually (training and testing on the same language $A$) or cross-lingually (training on language $A$ and testing on a different language $B$).
The regularization strength is selected from $\{10,100,1000\}$ by maximizing mean Spearman $\rho$ on the unseen split.

\paragraph{Evaluation Metric.}
Probe performance is measured using Spearman rank correlation ($\rho$) between predicted and ground-truth difficulty scores on held-out test problems.
This metric captures ordinal agreement and is robust to scale differences across probes.

\begin{table*}[t]
\centering
\footnotesize
\setlength{\tabcolsep}{3.8pt}
\renewcommand{\arraystretch}{1.15}
\begin{tabular}{l c c c c c c c}
\toprule
\multirow{2}{*}{\textbf{Model}} &
\multirow{2}{*}{\textbf{\shortstack{Num.\\[-0.2em]Layers}}} &
\multicolumn{2}{c}{\textbf{\shortstack{Probe performance\\[-0.2em]Spearman $\rho$ (mean $\pm$ std)}}} &
\multicolumn{2}{c}{\textbf{\shortstack{Optimal layer\\[-0.2em](mean peak layer)}}} &
\multirow{2}{*}{\textbf{\shortstack{$\Delta$ \\ Transfer drop\\[-0.2em](Diag$\rightarrow$Off-Diag)}}} &
\multirow{2}{*}{\textbf{\shortstack{$\Delta$ \\ In-lang drop\\[-0.2em](Off-Diag$\rightarrow$Diag)}}} \\
\cline{3-4}\cline{5-6}\\[-2ex]
&
&
\textbf{\shortstack{Same-lang\\[-0.2em](Diag.)}} &
\textbf{\shortstack{Cross-lang\\[-0.2em](Off-diag.)}} &
\textbf{\shortstack{Same-lang\\[-0.2em](Diag.)}} &
\textbf{\shortstack{Cross-lang\\[-0.2em](Off-diag.)}} &
&
\\
\midrule

\texttt{LLaMA-3.1-8B} &
32 &
$0.822 \pm 0.019$ &
$0.783 \pm 0.024$ &
$30.05$ &
$15.67$ &
$0.177$ &
$0.014$ \\ 

\texttt{LLaMA-3.2-3B} &
28 &
$0.805 \pm 0.022$ &
$0.771 \pm 0.027$ &
$22.62$ &
$9.07$ &
$0.192$ &
$0.010$ \\ 

\texttt{LLaMA-3.2-1B} &
16 &
$0.797 \pm 0.021$ &
$0.761 \pm 0.028$ &
$12.05$ &
$6.11$ &
$0.055$ &
$0.007$ \\  

\texttt{Qwen3-8B} &
36 &
$0.849 \pm 0.019$ &
$0.783 \pm 0.034$ &
$14.29$ &
$8.59$ &
$0.114$ &
$0.015$ \\ 

\bottomrule
\end{tabular}
\caption{
\textbf{Cross-lingual difficulty probing summary across models.}
Performance reports Spearman $\rho$ (mean $\pm$ std) over language pairs.
Across all models, cross-lingual (off-diagonal) performance is statistically lower than same-language (diagonal) performance under a paired significance test ($p < 10^{-3}$).
Optimal layer denotes the mean layer index achieving peak $\rho$ in each regime.
\textbf{Transfer drop (Diag$\rightarrow$Off-Diag)} measures the loss in cross-lingual performance when probes are fixed at the diagonal-optimal layer.
\textbf{In-lang drop (Off-Diag$\rightarrow$Diag)} measures the loss in same-language performance when probes are fixed at the transfer-optimal layer.
}
\label{tab:crosslingual-difficulty-summary}
\end{table*}

\section{Results}
\label{sec:results}

We present results for \texttt{LLaMA-3.1-8B} as a representative model; all trends described below are consistent across \texttt{LLaMA-3.2} (1B, 3B) and \texttt{Qwen3-8B}, with full results reported in Appendix~\ref{a:additional_results}.

\subsection{Depth-Dependent Structure of Difficulty Representations}
\label{sec:depth_divergence}

We first investigate whether problem difficulty is encoded in a language-specific manner or whether it corresponds to a shared internal representation across languages.

Figure~\ref{fig:spearman_matrices_Llama3_1} (left) summarizes cross-lingual probing performance. Each cell reports the \emph{maximum} Spearman correlation achieved across all layers for a given training–testing language pair. Diagonal entries correspond to same-language probing, while off-diagonal entries reflect cross-lingual transfer. Rows therefore indicate how well a probe trained on a given language generalizes to all others.

Overall, we observe uniformly high correlations across language pairs, including transfers between typologically distant and low-resource languages, indicating that relative difficulty rankings are largely preserved across languages.

Figure~\ref{fig:spearman_matrices_Llama3_1} (right) reveals a systematic difference in \emph{where} these correlations occur. For \texttt{LLaMA-3.1-8B}, probes trained and tested on the same language (i.e. diagonal scores) consistently peak in later layers (around layer~30), whereas cross-lingual transfer peaks much earlier (around layer~14). This pattern is highly stable across languages: diagonal cells concentrate tightly around a single later (deep) layer, while off-diagonal cells concentrate around earlier layers.

Table~\ref{tab:crosslingual-difficulty-summary} quantifies this separation. For \texttt{LLaMA-3.1-8B}, the mean optimal layer for same-language probing is 30.05, compared to 15.67 for cross-lingual transfer. The same depth separation appears across all evaluated models, with the absolute layer indices shifting according to model depth (Appendix~\ref{a:additional_results}).

Taken together, these results indicate a clear depth-dependent organization of difficulty representations: an earlier, shallow layer representation and a later, deeper representation that is optimized for language-specific performance.

\subsection{Cross-Lingual Generalization }

To further study the consequences of this depth divergence, we explicitly compare probe performance at layers optimized for same-language versus cross-lingual evaluation. We hypothesized that \emph{diagonal-optimal} layers are maximizing language specific performance (i.e. probes trained on those layers learn language specific features) while \emph{off-diagonal-optimal} layers are optimized for cross-lingual representation (i.e. represent features in a language-agnostic manner). 
The final two columns of Table~\ref{tab:crosslingual-difficulty-summary} report the performance impact of evaluating probes at these respective layer choices on same-language and cross-lingual scenarios (see Appendix~\ref{a:cross_lingual_details} for details).

When probes are evaluated cross-lingually at the diagonal-optimal layer, performance drops substantially. For \texttt{LLaMA-3.1-8B}, fixing probes at the diagonal optimal layer leads to a mean reduction of 0.177 Spearman~$\rho$ under transfer. This sharp decline indicates that deeper layers, while highly predictive within-language, encode problem-difficulty in a manner that does not align well across languages.

In contrast, evaluating probes monolingually at the off-diagonal-optimal layer results in only a negligible loss in same-language performance (0.014 Spearman~$\rho$ for \texttt{LLaMA-3.1-8B}). Thus, the layer that best supports cross-lingual transfer (shallow layer) remains near-optimal for the source language itself.

Together, these findings indicate that problem difficulty is organized around a shared, language-independent direction in activation space that emerges at shallow layers. Deeper layers refine this signal in a language-specific way, improving within-language accuracy at the expense of cross-lingual alignment. This depth-dependent trade-off explains why cross-lingual generalization peaks earlier in the network while same-language performance continues to improve at later layers.

\section*{Conclusion}
By probing residual activations across 21 languages, we find that LLMs encode problem difficulty in a depth-dependent way: an early, shared representation supports strong cross-lingual transfer, while deeper layers refine difficulty in a language-specific manner, improving monolingual accuracy but reducing alignment across languages. This shared difficulty direction emerges early and remains stable even for low-resource languages, indicating that LLMs form a language-agnostic estimate of difficulty before specializing it to individual languages.

These findings extend prior work by \cite{lugoloobi2025llms} in two important ways. First, they demonstrate that the difficulty signal identified in English is not an artifact of language, but instead reflects a genuinely multilingual internal property. Second, they suggest that the mechanistic role of difficulty, previously shown to support steering and hallucination reduction, originates in a shared representational subspace that precedes language-specific reasoning. This supports viewing difficulty as a high-level internal signal of the model, rather than a byproduct of surface-level language features.

From a practical perspective, our results point to new opportunities for multilingual systems, especially in low-resource settings. Lightweight difficulty predictors trained on shallow activations can be deployed without per-language tuning, enabling language-agnostic routing, curriculum design, or compute-aware inference even where labeled data are scarce. 

\section*{Limitations}

While we show that problem difficulty is encoded in a shared, cross-lingual subspace, our analysis is confined to mathematical problem solving using the AMC subset of Easy2Hard. This domain offers the advantage of well-calibrated, human-derived difficulty labels, but it represents a narrow slice of model behavior. It therefore remains unclear whether the same cross-lingual geometry of difficulty extends to domains where difficulty is more subjective or context-dependent, such as commonsense reasoning, programming, or open-ended generation.

Additionally, our empirical evaluation is limited to a small set of instruction-tuned, decoder-only language models. Although the observed trends are consistent across them, it is not yet clear whether the same representational structure is present in all model.

Finally, our conclusions are based exclusively on probing analyses. Probing establishes the presence and cross-lingual alignment of difficulty-related signals, but does not by itself demonstrate that these signals play a causal role in shaping model behavior during inference. Prior work provides such causal evidence in the English setting \cite{lugoloobi2025llms}, showing that intervening along a learned difficulty direction can steer model behavior and reduce hallucinations. Whether analogous interventions transfer across languages—for example, by steering a model using difficulty vectors learned in one language and applied to another—remains an open question, which we leave to future work.

\bibliography{custom}

\appendix

\section{API Cost Analysis}
\label{a:costs}
To construct the multilingual version of the Easy2Hard AMC benchmark, we translated all English problems into 20 additional languages using the \texttt{gpt-5.1} API.
Each translation call used a fixed prompt in which, for each problem–language pair, the placeholders \textbf{\texttt{[TARGET LANGUAGE]}} and \textbf{\texttt{[PROBLEM\_TEXT]}} were replaced with the corresponding language identifier and problem content, respectively (see prompt box).
The benchmark contains 3,975 unique problems, yielding a total of 79,500 translations.

Using the official API pricing at the time of experimentation (input: \$0.625/M tokens, output: \$5.00/M tokens, cached input: \$0.0625/M tokens.
Overall, the full multilingual benchmark was produced for less than \$50 USD.

\begin{tcolorbox}[
  colback=gray!10!white,
  colframe=gray!75!black,
  title=\textbf{Prompt for Math Problem Translation},
  fonttitle=\bfseries,
  boxrule=1pt,
  width=\linewidth
]
\small
\textbf{<SYSTEM PROMPT>} 

You are a helpful assistant that translates math problems accurately.
\medskip
\medskip

\textbf{<USER PROMPT>}

Translate the following math problem into \textbf{[TARGET LANGUAGE]}.
\medskip

Problem:
\textbf{[PROBLEM\_TEXT]}
\medskip

Translation:
\end{tcolorbox}

\section{Translation Quality Assessment}
\label{a:t_quality}

\begin{table}[b!]
\centering
\small
\caption{\textbf{Average COMET-Kiwi scores for translated problems by target language.}
Higher scores indicate stronger semantic adequacy with respect to the source language (English).
}
\label{tab:comet-kiwi}
\begin{tabular}{lc}
\toprule
\textbf{Language} & \textbf{COMET-Kiwi Score} \\
\midrule
Italian      & 0.8250 \\
Japanese     & 0.8232 \\
Dutch        & 0.8232 \\
French       & 0.8218 \\
Spanish      & 0.8183 \\
Turkish      & 0.8106 \\
Vietnamese   & 0.8101 \\
Russian      & 0.8069 \\
Portuguese   & 0.8052 \\
Korean       & 0.8050 \\
Greek        & 0.8045 \\
Chinese      & 0.8026 \\
Bengali      & 0.8019 \\
German       & 0.8002 \\
Thai         & 0.7906 \\
Polish       & 0.7894 \\
Hebrew       & 0.7681 \\
Arabic       & 0.7576 \\
Telugu       & 0.7553 \\
Swahili      & 0.6134 \\
\bottomrule
\end{tabular}
\end{table}

To assess the semantic adequacy of the automatically generated translations used in our multilingual benchmark, we employ COMET-Kiwi, a reference-free machine translation quality estimation metric introduced by \citet{rei2022cometkiwi}. 
Unlike traditional n-gram--based metrics (e.g., BLEU), COMET-Kiwi estimates translation quality by predicting human adequacy judgments from multilingual neural representations, without requiring reference translations \cite{ju2024large}.

Table~\ref{tab:comet-kiwi} reports the average COMET-Kiwi score for each target language. Scores are consistently high across the majority of languages, with most values exceeding $0.75$, indicating strong relative semantic adequacy with respect to the English source. 
High-resource European languages (e.g., Italian, French, Spanish, German) achieve the highest scores, while typologically distant and lower-resource languages exhibit more modest degradation, most notably Swahili.

Importantly, COMET-Kiwi scores are not calibrated to absolute quality thresholds and are intended to be interpreted comparatively rather than as guarantees of human-level translation quality.
Accordingly, we treat these results as evidence against severe semantic distortion rather than as a definitive certification of translation correctness.

Crucially, translation adequacy is further validated indirectly through downstream probing behavior. If translation quality were poor or systematically distorted problem semantics, difficulty probes trained on one language would fail to transfer to others.
Instead, we observe strong cross-lingual probe transfer across all languages considered, including those with lower COMET-Kiwi scores, indicating that the translated problems preserve the underlying difficulty signal required for our analysis.
This task-level invariance provides complementary evidence that residual translation noise does not materially affect our main findings.

\section{Additional Results}
\label{a:additional_results}

This appendix reports supplementary analyses for \texttt{LLaMA-3.2-3B}, \texttt{LLaMA-3.2-1B}, and \texttt{Qwen3-8B}, extending the main results presented for \texttt{LLaMA-3.1-8B}. As summarized in Table~\ref{tab:crosslingual-difficulty-summary}, all three models exhibit the same qualitative trends discussed in Section~\ref{sec:results}; the figures here provide a more fine-grained, model-specific view.

\begin{figure*}[t!]
    \centering
    \begin{subfigure}[t]{0.85\linewidth}
        \centering
        \includegraphics[width=\linewidth]{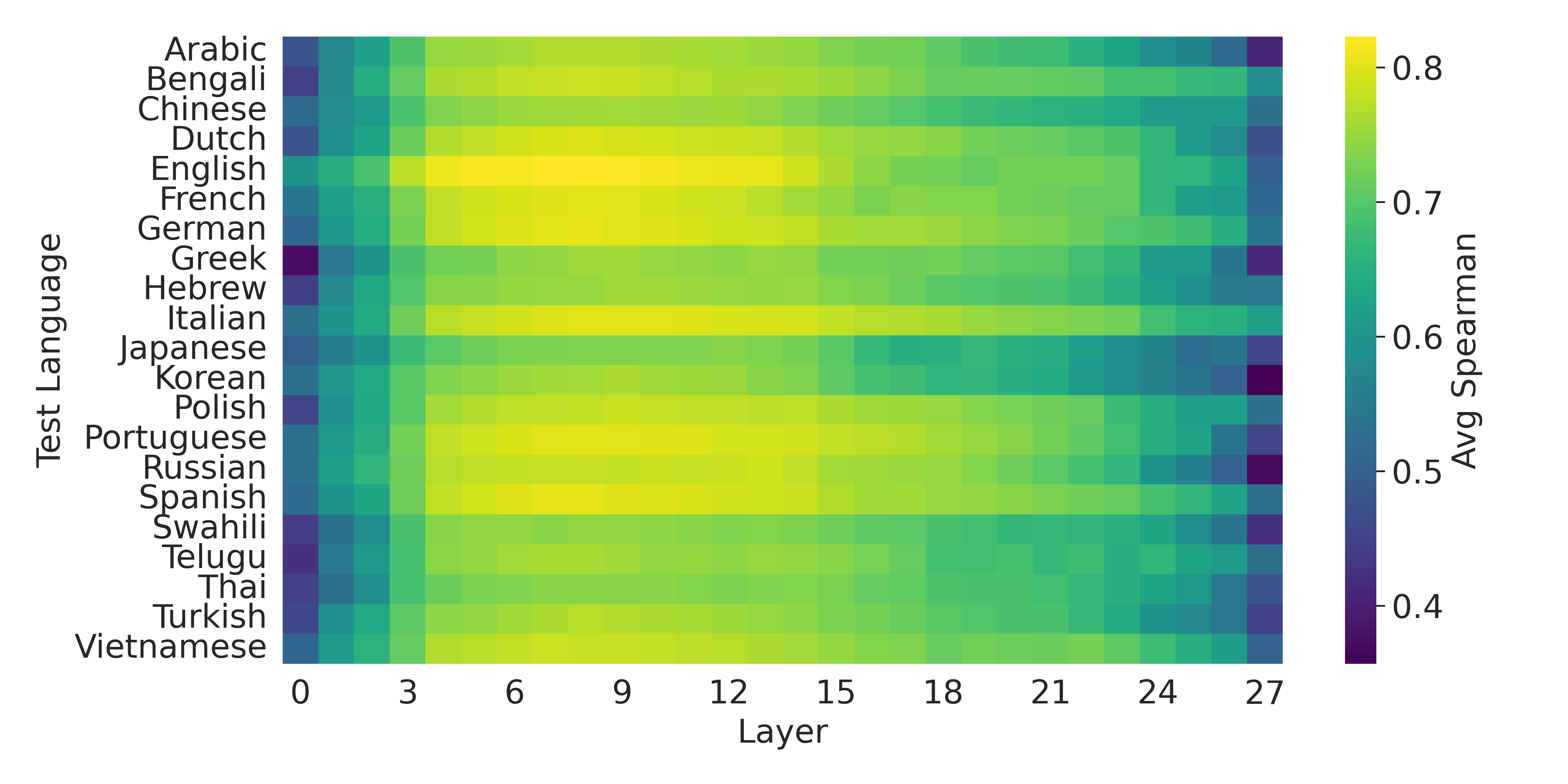}
        \caption{Llama3.2\_3B}
        \label{fig:fig1a}
    \end{subfigure}

    \vspace{0.6em}

    \begin{subfigure}[t]{0.85\linewidth}
        \centering
        \includegraphics[width=\linewidth]{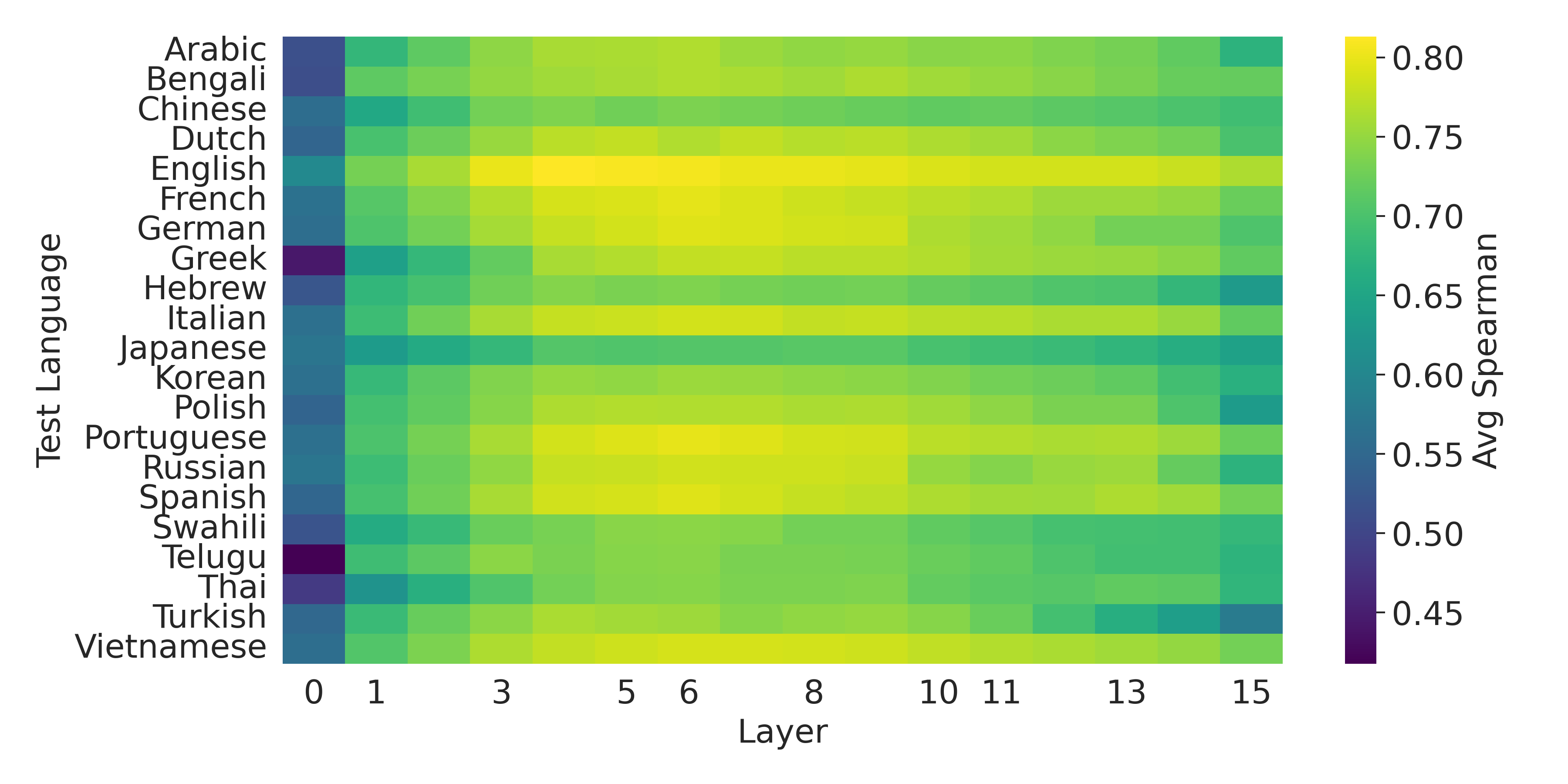}
        \caption{Llama3.2\_1B}
        \label{fig:fig1b}
    \end{subfigure}

    \vspace{0.6em}

    \begin{subfigure}[t]{0.85\linewidth}
        \centering
        \includegraphics[width=\linewidth]{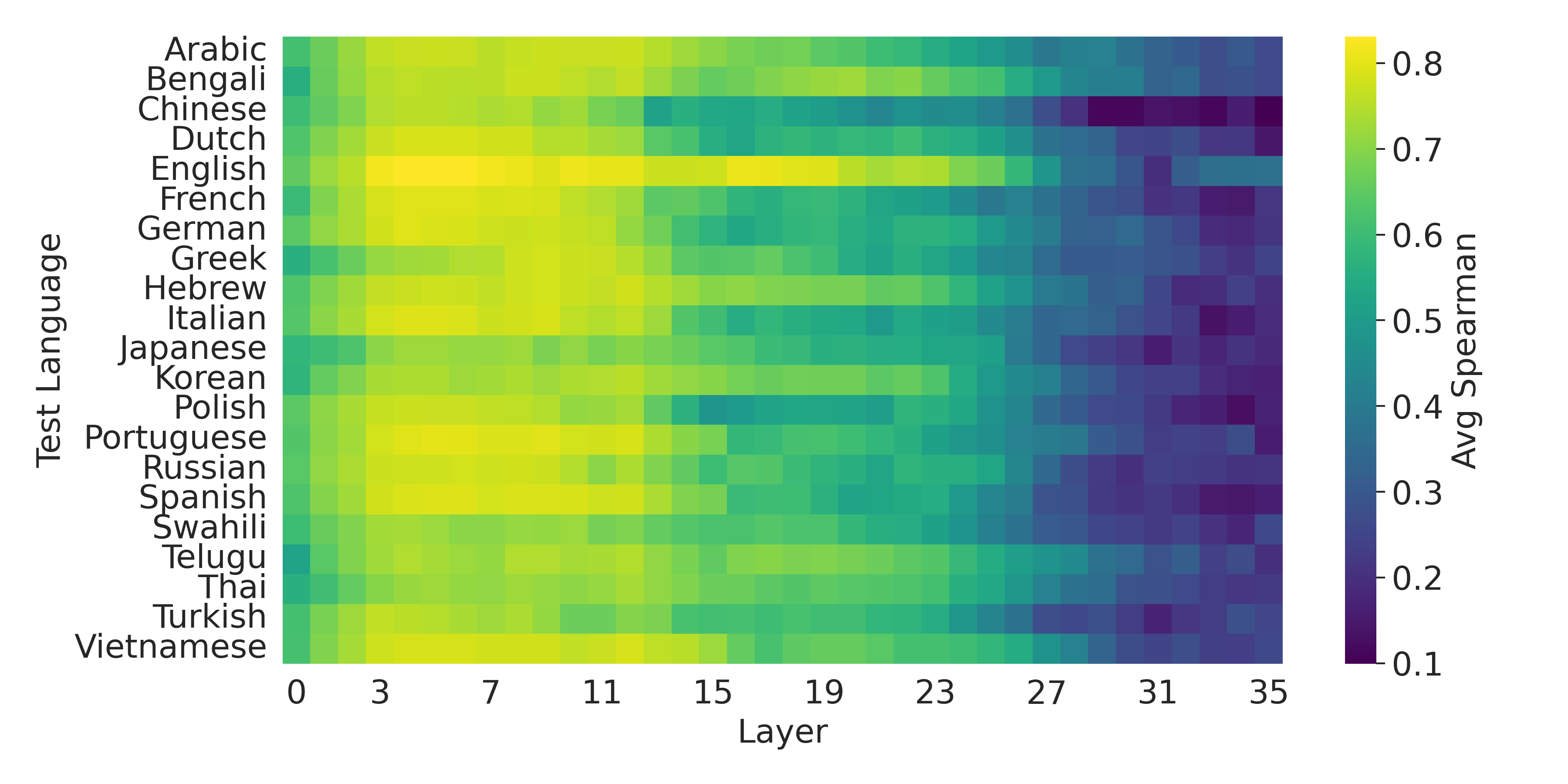}
        \caption{Qwen3}
        \label{fig:fig1c}
    \end{subfigure}

    \caption{Layer-wise cross-lingual difficulty probing for additional models. Same setup as Figure~\ref{fig:layerwise-transfer}, but for LLaMA-3.2-3B, LLaMA-3.2-1B, and Qwen3-8B. Heatmaps report, for each test language and transformer layer, the average Spearman correlation between predicted and ground-truth difficulty, averaged over probes trained on all other languages. As in Figure~\ref{fig:layerwise-transfer}, cross-lingual performance peaks in early-to-middle layers across models.}
    \label{fig:additional-layerwise}
\end{figure*}

\paragraph{Layer-wise cross-lingual performance.}
Figure~\ref{fig:additional-layerwise} reports the average cross-lingual Spearman correlation as a function of layer and test language. For both \texttt{LLaMA-3.2} variants, probe performance peaks in early-to-middle layers and degrades toward the top of the network, closely mirroring the depth-dependent divergence observed for \texttt{LLaMA-3.1-8B}. \texttt{Qwen3-8B} shows a similar early peak, followed by a sharper decline in later layers, consistent with its larger separation between same-language and cross-lingual optimal layers reported in Table~\ref{tab:crosslingual-difficulty-summary}. Overall, these results reinforce the conclusion that the most transferable difficulty signal emerges before language-specific processing dominates deeper layers.

\begin{figure*}[t!]
    \centering
    % Row 1: Llama3.2_3B
    \begin{subfigure}[t]{0.48\linewidth}
        \centering
        \includegraphics[width=\linewidth]{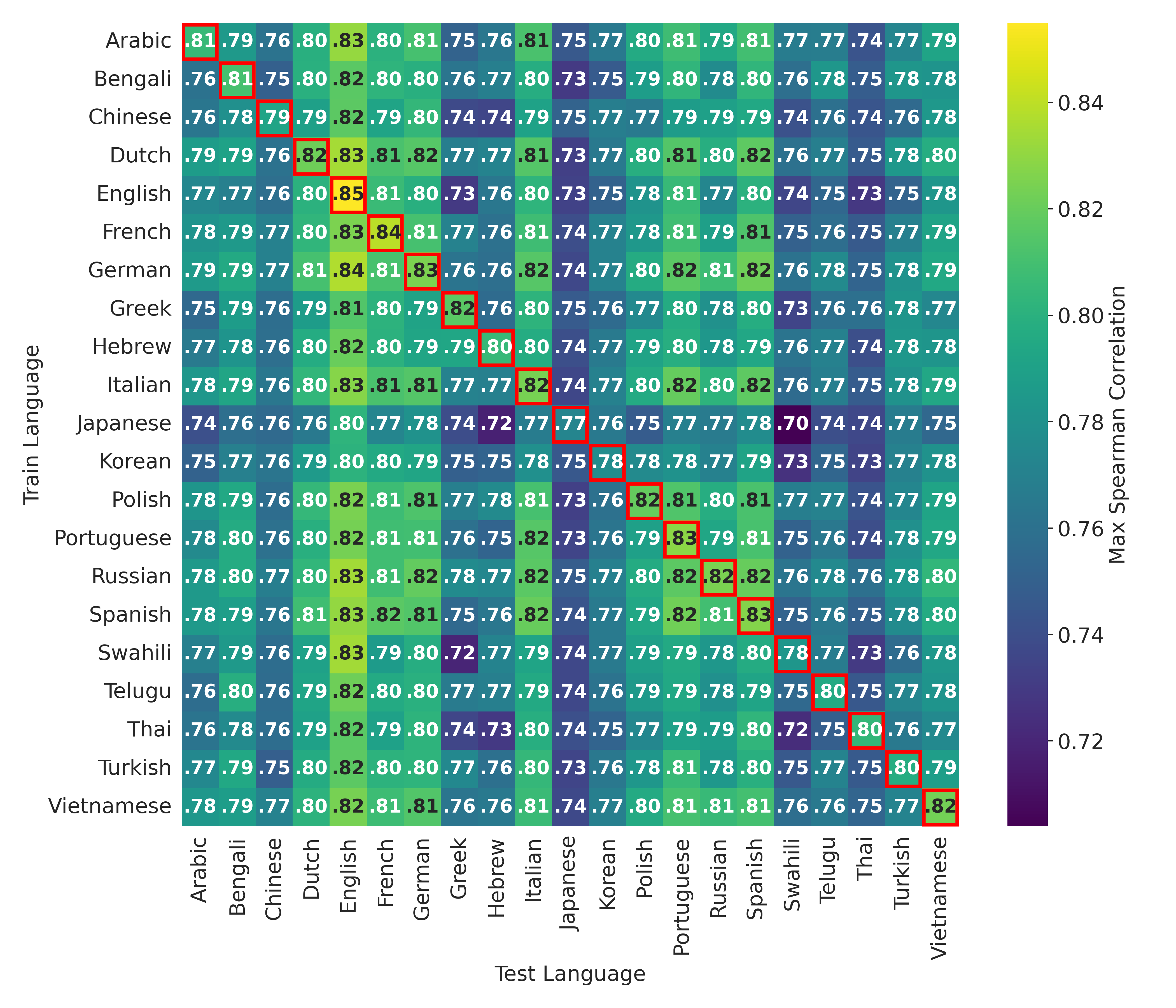}
        \caption{Llama3.2\_3B (problem-wise)}
    \end{subfigure}
    \hfill
    \begin{subfigure}[t]{0.48\linewidth}
        \centering
        \includegraphics[width=\linewidth]{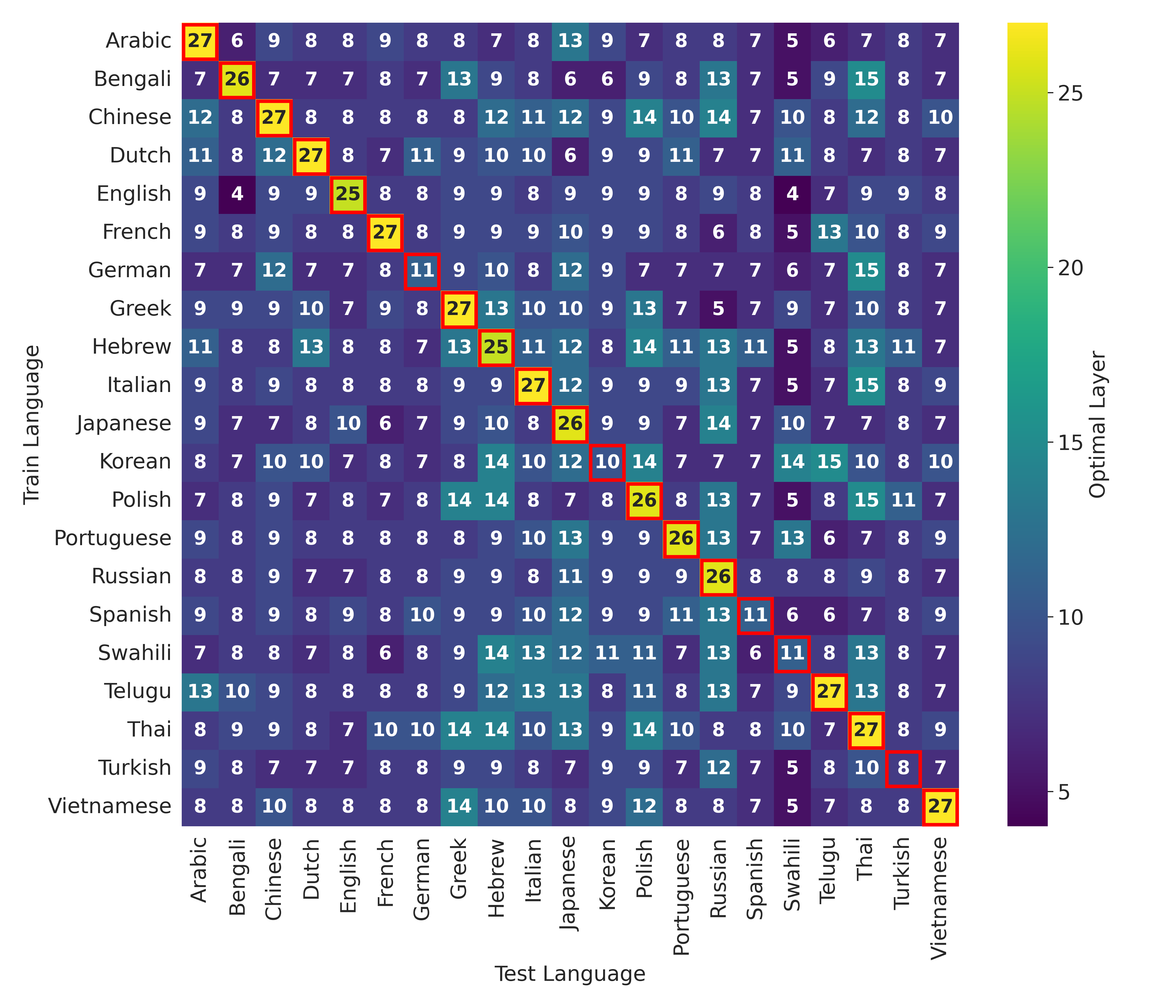}
        \caption{Llama3.2\_3B (layer-wise)}
    \end{subfigure}

    \vspace{0.8em}

    % Row 2: Llama3.2_1B
    \begin{subfigure}[t]{0.48\linewidth}
        \centering
        \includegraphics[width=\linewidth]{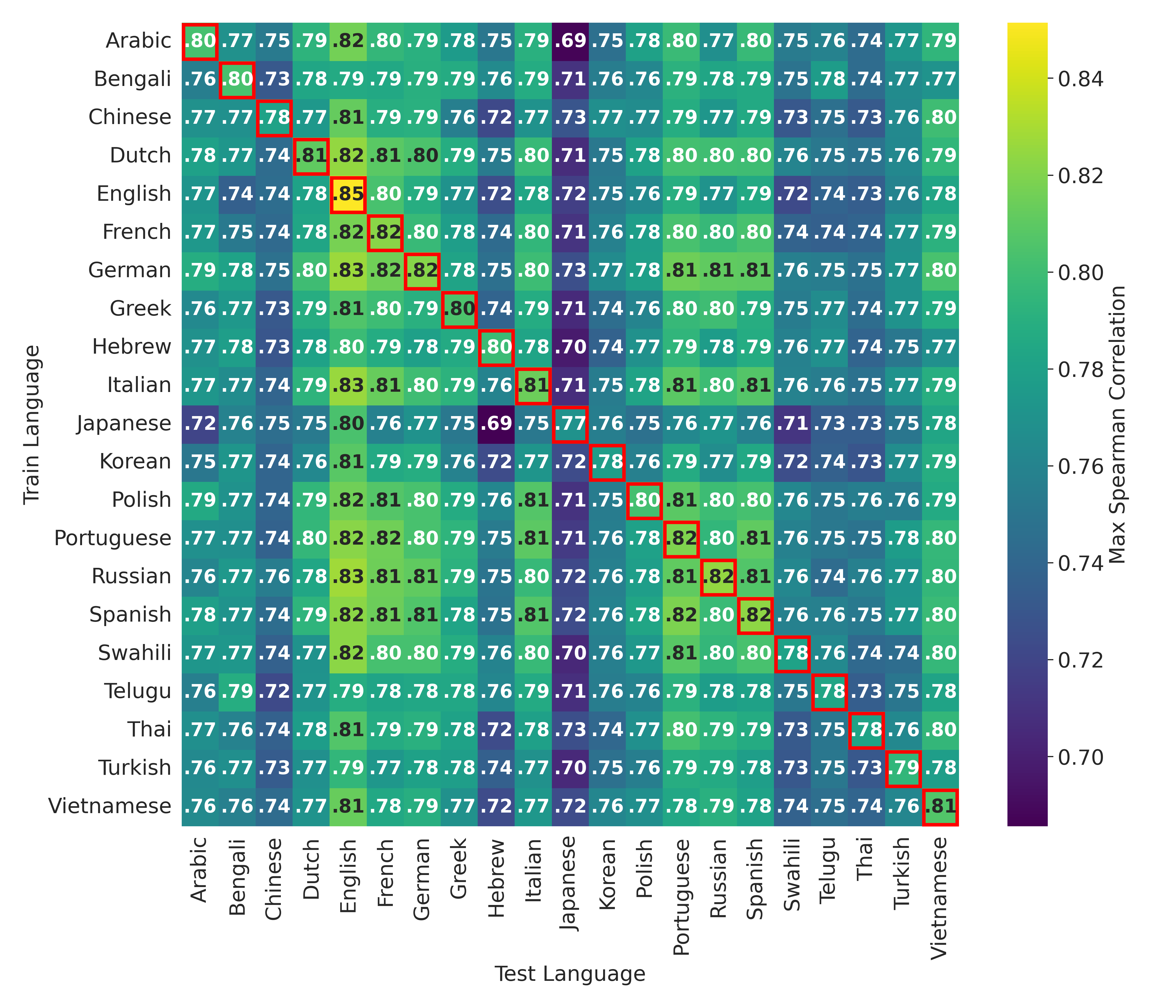}
        \caption{Llama3.2\_1B (problem-wise)}
    \end{subfigure}
    \hfill
    \begin{subfigure}[t]{0.48\linewidth}
        \centering
        \includegraphics[width=\linewidth]{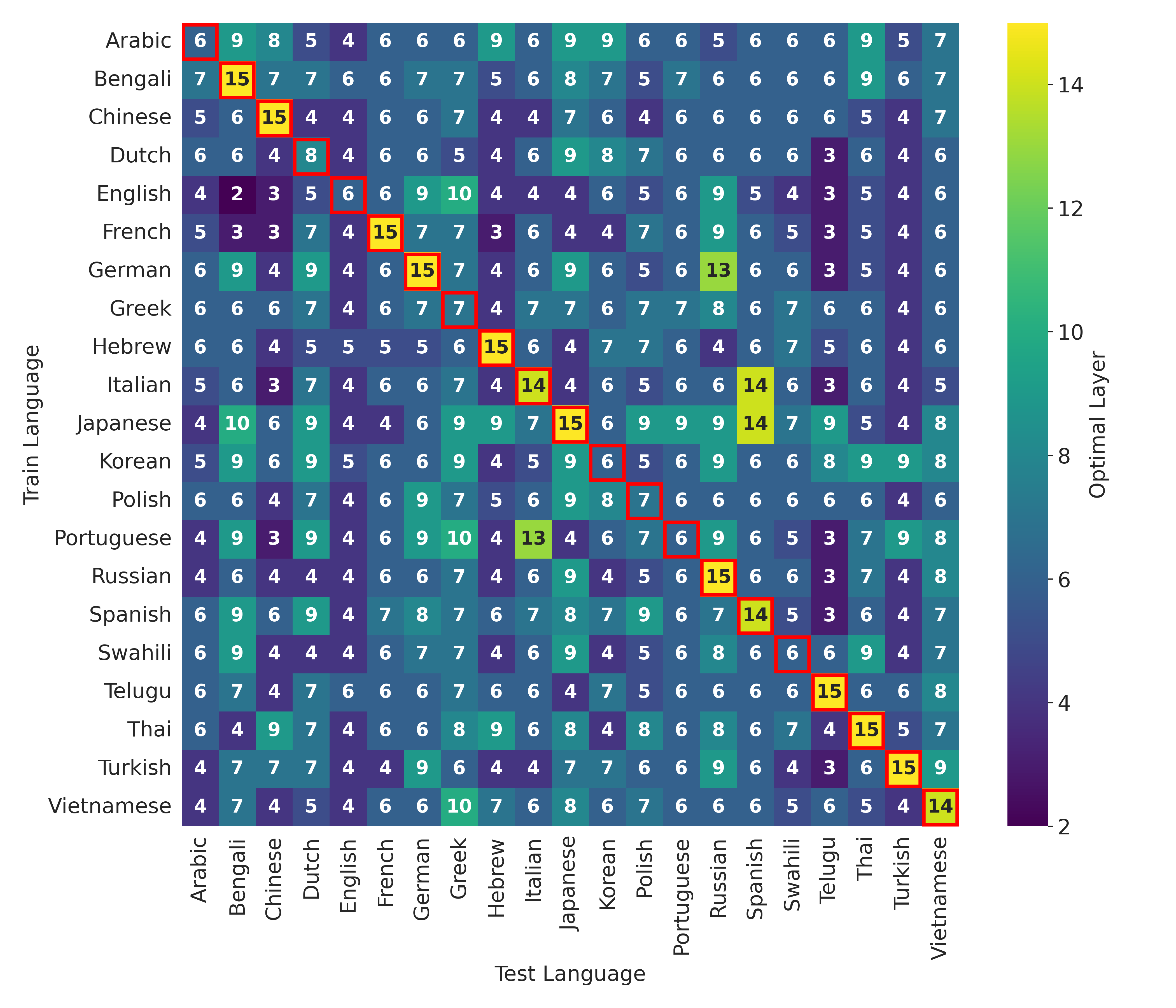}
        \caption{Llama3.2\_1B (layer-wise)}
    \end{subfigure}

    \vspace{0.8em}

    % Row 3: Qwen3
    \begin{subfigure}[t]{0.48\linewidth}
        \centering
        \includegraphics[width=\linewidth]{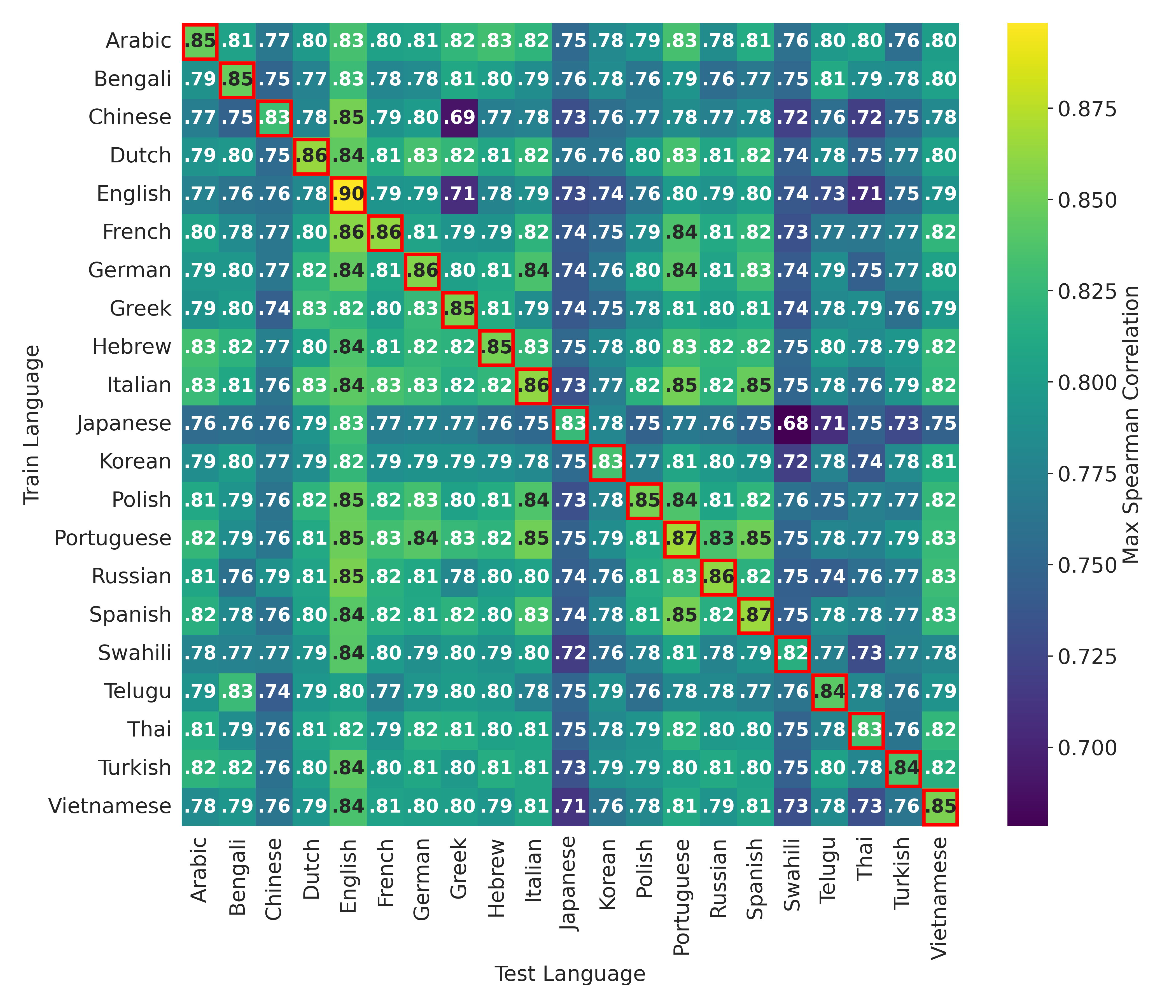}
        \caption{Qwen3 (problem-wise)}
    \end{subfigure}
    \hfill
    \begin{subfigure}[t]{0.48\linewidth}
        \centering
        \includegraphics[width=\linewidth]{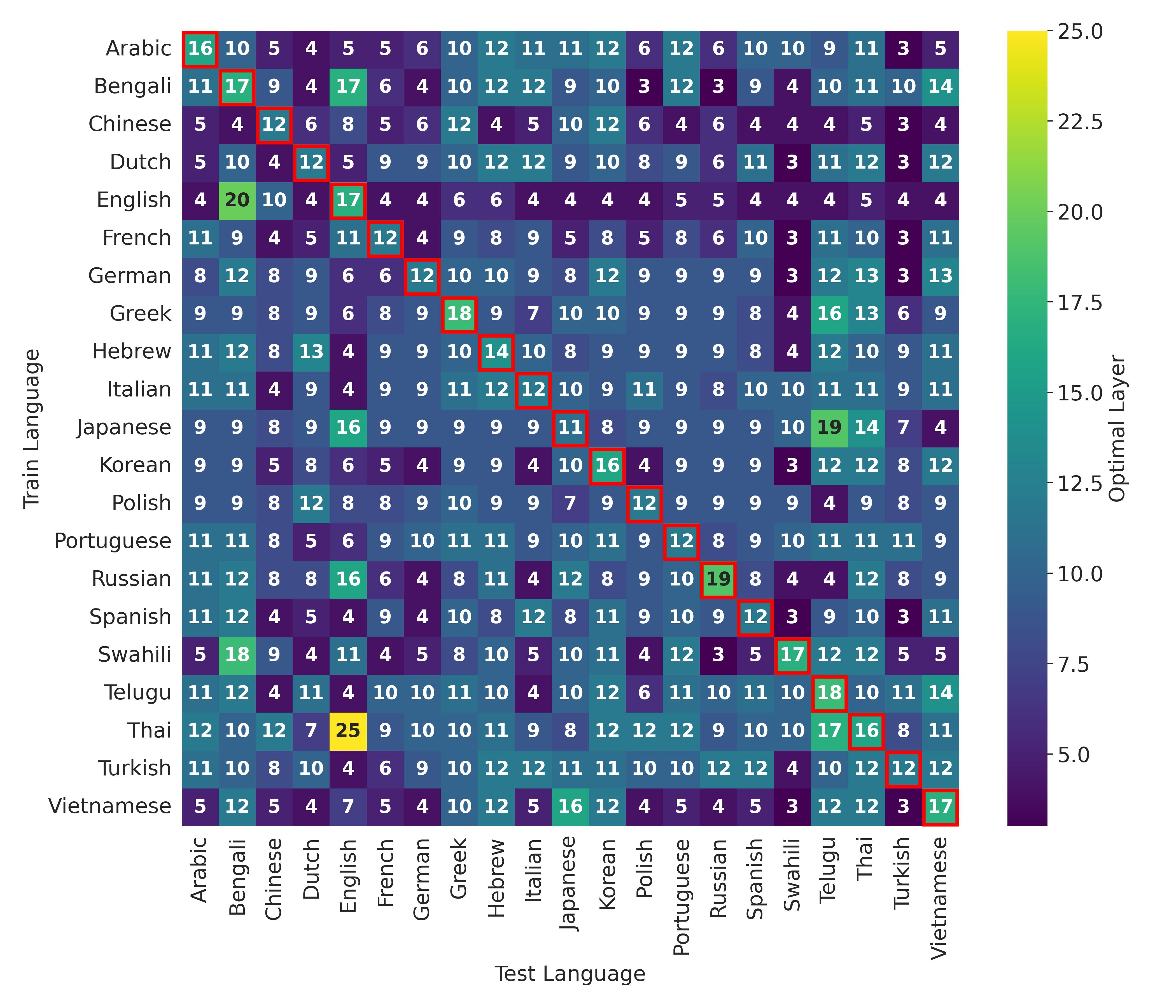}
        \caption{Qwen3 (layer-wise)}
    \end{subfigure}

    \caption{Cross-lingual structure of difficulty representations for additional models. Same analysis as Figure~\ref{fig:spearman_matrices_Llama3_1} but for LLaMA-3.2-3B, LLaMA-3.2-1B, and Qwen3-8B. Left: maximum Spearman $\rho$ for each training–testing language pair at its optimal layer. Right: corresponding layer indices achieving peak performance. The same pattern holds across models, with cross-lingual optima concentrated at earlier layers than monolingual optima.}
    \label{fig:additional-matrices}
\end{figure*}

\paragraph{Cross-lingual consistency across languages.}
Figure~\ref{fig:additional-matrices} presents full Spearman correlation matrices for each model. The problem-wise matrices (left column) are uniformly high across language pairs, indicating that relative difficulty rankings are preserved across translations. The layer-wise matrices (right column) show that cross-lingual alignment concentrates within a narrow band of earlier layers, while optimal layers diverge in deeper regions. This effect is most pronounced for \texttt{LLaMA-3.2-1B}, aligning with Table~\ref{tab:crosslingual-difficulty-summary}, which shows that reduced model capacity shifts transfer-optimal layers earlier in the network.

\section{Cross-Lingual Generalization Analysis}
\label{a:cross_lingual_details}

This appendix details the procedure used to compute the \emph{Transfer drop} and \emph{In-language drop} reported in the final two columns of Table~\ref{tab:crosslingual-difficulty-summary}.
\begin{enumerate}
    \item \textbf{Layer-wise probing.}
    For each training–testing language pair $(A,B)$, linear difficulty probes are evaluated at every transformer layer.
    Performance is measured using Spearman $\rho$ on held-out test problems, yielding a layer-wise performance profile for each pair.

    \item \textbf{Diagonal-optimal layers.}
    For each language $A$, the \emph{diagonal-optimal layer} is defined as the layer that maximizes performance when training and testing on the same language $(A,A)$.
    This layer corresponds to the depth at which same-language difficulty encoding is strongest.

    \item \textbf{Transfer drop (Diag$\rightarrow$Off-Diag).}
    For a fixed training language $A$ and each target language $B \neq A$:
    \begin{itemize}
        \item compute the best achievable cross-lingual performance for $(A,B)$ across all layers;
        \item compute the cross-lingual performance obtained when evaluating at $A$’s diagonal-optimal layer;
        \item take the difference between the two.
    \end{itemize}
    These differences are averaged across all $B \neq A$ and then across all $A$ to obtain the mean \emph{Transfer drop}, measuring how much cross-lingual performance is lost when using same-language–optimal layers.

    \item \textbf{Transfer-optimal layers.}
    For each training language $A$:
    \begin{itemize}
        \item identify, for each target language $B \neq A$, the layer that maximizes performance for $(A,B)$;
        \item take the statistical mode of these layers to obtain a single \emph{transfer-optimal layer}.
    \end{itemize}

    \item \textbf{In-language drop (Off-Diag$\rightarrow$Diag).}
    For each language $A$:
    \begin{itemize}
        \item compute the best same-language performance across all layers;
        \item compute the same-language performance at the transfer-optimal layer;
        \item take the difference between the two.
    \end{itemize}
    These differences are averaged across languages to yield the mean \emph{In-language drop}, quantifying the cost of fixing probes at transfer-optimal layers for monolingual evaluation.
\end{enumerate}

\clearpage

\end{document}